\documentclass[conference]{IEEEtran}
\IEEEoverridecommandlockouts
\usepackage{cite}
\usepackage{amsmath,amssymb,amsfonts}
\usepackage{algorithmic}
\usepackage{graphicx}
\usepackage{textcomp}
\usepackage{xcolor}
\usepackage{multirow}
\usepackage{booktabs}
\usepackage{authblk}
\def\BibTeX{{\rm B\kern-.05em{\sc i\kern-.025em b}\kern-.08em
    T\kern-.1667em\lower.7ex\hbox{E}\kern-.125emX}}
\begin{document}

\title{InstructVid2Vid: Controllable Video Editing with Natural Language Instructions
\thanks{This work was supported by the National Key Research and Development Project of China (2022ZD0160101), Key Research and Development Projects in Zhejiang Province (No. 2024C01106), and Research funding from FinVolution Group.).

$^\dag$Corresponding author}

}


\author[a]{Bosheng Qin}
\author[a]{Juncheng Li}
\author[a]{Siliang Tang}
\author[b]{Tat-Seng Chua}
\author[a]{Yueting Zhuang$^{\dag,}$}
\affil[a]{College of Computer Science and Technology, Zhejiang University}
\affil[a]{$\{$bsqin, junchengli, siliang, yzhuang$\}$@zju.edu.cn}
\affil[b]{School of Computing, National University of Singapore}
\affil[b]{dcscts@nus.edu.sg}

\maketitle

\begin{abstract}
We introduce \textbf{InstructVid2Vid}, an end-to-end diffusion-based methodology for video editing guided by human language instructions. Our approach empowers video manipulation guided by natural language directives, eliminating the need for per-example fine-tuning or inversion. The proposed InstructVid2Vid model modifies a pretrained image generation model, Stable Diffusion, to generate a time-dependent sequence of video frames. By harnessing the collective intelligence of disparate models, we engineer a training dataset rich in video-instruction triplets, which is a more cost-efficient alternative to collecting data in real-world scenarios. To enhance the coherence between successive frames within the generated videos, we propose the Inter-Frames Consistency Loss and incorporate it during the training process. With multimodal classifier-free guidance during the inference stage, the generated videos is able to resonate with both the input video and the accompanying instructions. Experimental results demonstrate that InstructVid2Vid is capable of generating high-quality, temporally coherent videos and performing diverse edits, including attribute editing, background changes, and style transfer. These results underscore the versatility and effectiveness of our proposed method.
\end{abstract}


\section{Introduction}
\label{sec:intro}

Diffusion models have displayed impressive efficacy in image editing tasks \cite{RN10108}. Recent endeavors aim to extend this success from images to videos, exemplified by Dreamix \cite{RN10134}, Video-P2P \cite{RN10116}, and Tune-a-Video \cite{RN10117}. These approaches necessitate a fine-tuning process for each video requiring editing. Initially, the video diffusion model is fine-tuned using the provided input video and a text description. This procedure establishes a correlation between specific text and the video content. Subsequently, during inference, modified versions of the input text are used to generate diverse edited renditions of the input video, yielding varied editing outcomes for the same original video. While effective, these methods demand exhaustive fine-tuning tailored to individual videos, substantially impeding editing efficiency.

To address the aforementioned problem, we propose InstructVid2Vid, a novel end-to-end video editing pipeline that edit the input video based on accepts human language instructions without requiring for any example-specific fine-tuning, extra images/videos, or comprehensive descriptions of the input/output videos. InstructVid2Vid generates the edited video based on the given input video and instructions, which edit the video with a conditional diffusion model. To effectively train InstructVid2Vid, we need to gather data that includes a variety of input videos, instructions, and corresponding output videos. It is crucial that the input and output videos demonstrate temporal consistency, and maintain consistency with the unedited subject, while accurately reflecting the modifications specified by the instructions. Unfortunately, there is no pre-existing dataset that readily provides the required data, and collecting video-instruction triplets in real-world situations can be expensive. Recent research suggests that the generative capabilities of a model can be enhanced by integrating multiple foundational models, offering a promising avenue for the development of a comprehensive solution \cite{RN10143, RN10108}. Based on the concept of model composition, we propose synthesizing video-instruction triplets by leveraging the combined use of ChatGPT \footnote{https://openai.com/blog/chatgpt}, video caption model \cite{RN10113}, and Tune-a-Video model \cite{RN10117}. The three models contain complementary knowledge between video and language, enabling us to build a more powerful pipeline for acquiring input videos, instructions, and edited videos triplets at a lower cost than collecting data in real-world scenarios. The resulting dataset will be used for training the InstructVid2Vid.

\begin{figure*}[t]
\centering
\includegraphics[width= \textwidth]{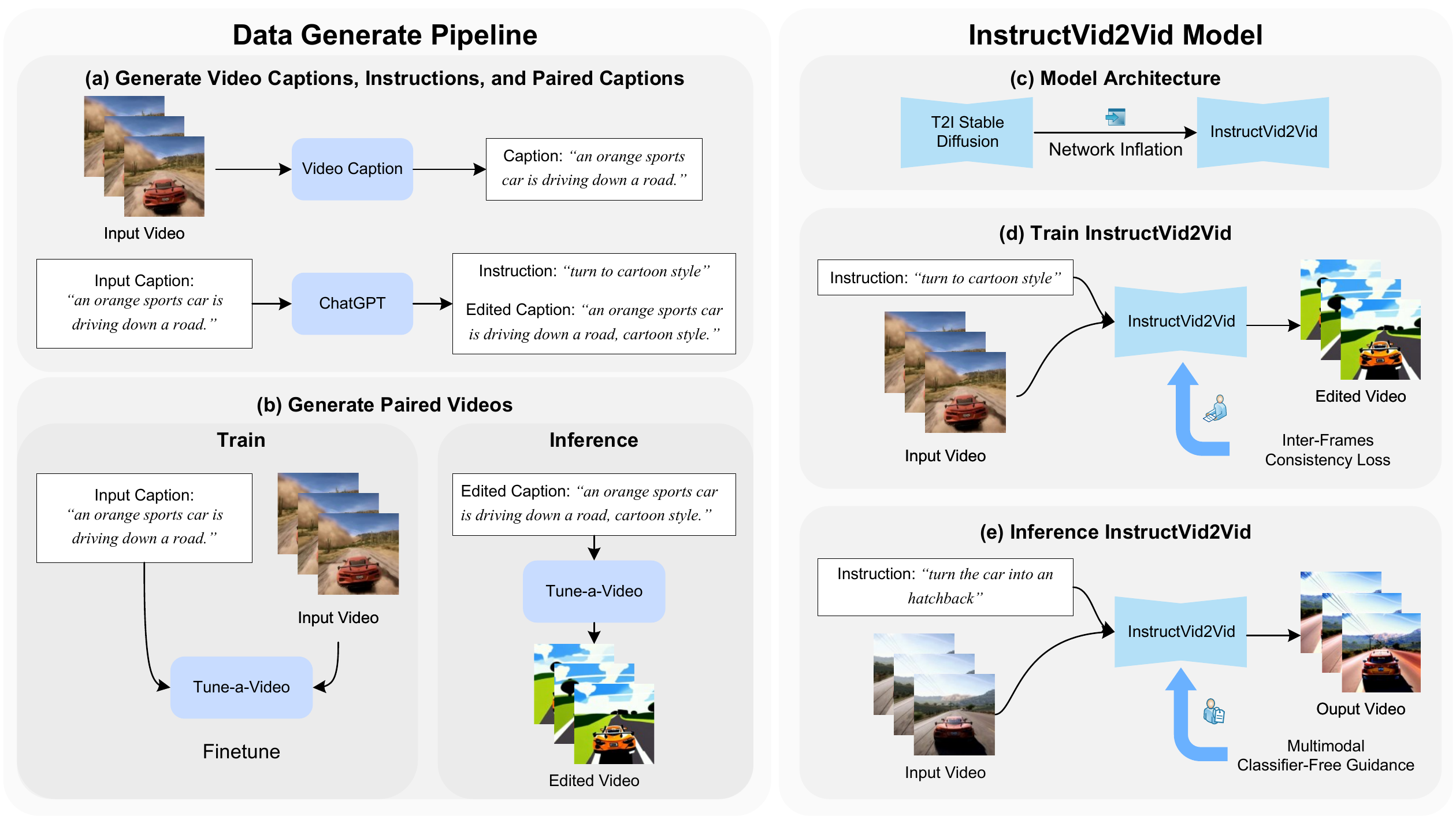}
\caption{Pipeline of InstructVid2Vid. (a) We first employ video caption model to generate video captions for real videos. Subsequently, we employ the prowess of ChatGPT to generate instructions and edited captions based on the video captions. (b) The synergy between input videos and captions fuels the fine-tuning of video diffusion model through the Tune-a-Video pipeline, and we obtain the edited video by inputting the edited captions into the fine-tuned video diffusion model. Finally, we acquire the input videos, instructions, and output video triplets. This dataset is used to train the (c) InstructVid2Vid model, where the denoising network is inherit and modified from T2I Stable Diffusion. (d) The training objective is to reconstruct a smooth, edited video with the proposed Inter-Frames Consistency Loss. (e) During inference, InstructVid2Vid edits input videos based on instructions with multimodal classifier-free guidance.}
\label{fig1}
\end{figure*}

\begin{table*}[t]
\caption{Examples of generated instructions and edited captions with video caption “a man in white is playing Tai Chi in the garden”.}
\label{tabel1}
\begin{center}
\resizebox{\textwidth}{!}{
\begin{tabular}{cccc}
\toprule
&Instruction& Edited Caption& Edit Type \\
\midrule 
\multirow{2}{*}{Handwirtten Example} &turn the man into a rabbit&a rabbit is playing Tai Chi in the garden& attribute modification\\
&turn to cartoon style&a man is playing Tai Chi in the garden, cartoon style& style transfer\\
\midrule 
\multirow{3}{*}{ChatGPT Generated }&move to desert ranch&a man in white is playing Tai Chi in the desert ranch & background change\\
&turn to charcoal pencil sketch style&a man in white is playing Tai Chi on a mountain, charcoal pencil sketch style& style transfer\\
&turn the man into a panda&a panda is playing Tai Chi in the garden& attribute modification\\
\bottomrule
\end{tabular}}
\end{center}
\end{table*}

During training, the InstructVid2Vid learns to reconstruct the ground truth edited video with the given instructions and input video. We propose the Inter-Frames Consistency Loss and apply it in training, which guides the model to generate video with consistency between adjacent frames. During inference, the model performs direct video editing in the forward pass in multimodal classifier-free guidance with text-video input, guiding the synthesized video to be more related to the input video and instructions. Experiments show that InstructVid2Vid is capable of generating high-quality videos with temporal coherence.

Our main contributions can be summarized as follows:

1)	We introduce InstructVid2Vid, an innovative, end-to-end video editing pipeline that can accomplish the video editing task without necessitating any example-specific fine-tuning or inversion processes for the provided input video.

2)	We put forward a novel pipeline for the synthesis of video-instruction triplets, to be employed as the training set for InstructVid2Vid. This process integrates a broad spectrum of knowledge and expertise sourced from different models. The resultant approach is markedly more cost-efficient than the traditional method of gathering data from real-world scenarios.

3)	We propose the integration of Inter-Frames Consistency Loss into the training procedure, which enhances the temporal consistency between adjacent frames in the edited videos.

4)	Experiment results demonstrate that InstructVid2Vid is capable of performing various types of video editing, including attribute editing, background alteration, and style transfer, resulting high-quality and temporally coherent edited videos.

\section{Related Work}
\subsection{Text Driven Video Editing}
Text-driven video editing has also piqued research interest \cite{RN10176, RN10179, RN10180}. Text2Live \cite{RN10112} aims to generate an edit layer that is composited over the input image or video with a given text prompt to augment the scene with visual effects. Recently, several video editing approaches have been proposed by fine-tuning the given images or videos with pretrained diffusion models. Dreamix \cite{RN10134} performs text-based motion and appearance editing through mixed video-image fine-tuning. Video-P2P \cite{RN10116} presents real-world video editing with cross-attention control using a decoupled-guidance strategy. The input video is inverted with the Text-to-Set model and optimized with a shared unconditional embedding. Turn-A-Video \cite{RN10117} generates videos from text prompts via one-shot fine-tuning of pretrained T2I diffusion models. Pix2Video \cite{RN10148} achieves video editing by DDIM inversion \cite{song2021denoising} and progressively propagate the changes to the intermedia features in self-attention. 

Distinct from previous works, our InstructVid2Vid model can achieve end-to-end video editing follow human language instructions without any example-specific fine-tuning or inversion for each video and does not require any additional input, such as user-drawn masks or extra images/videos.

\section{Method}
InstructVid2Vid is a supervised learning model that necessitates a training dataset. We will first introduce the pipeline to generate a training dataset for InstructVid2Vid. Then, We will present the model framework of InstructVid2Vid. Figure \ref{fig1} summarizes the pipeline of the proposed InstructVid2Vid.

\subsection{Dataset \label{Dataset Prepare}}

\subsubsection{Generate Video Captions, Instructions, and Edited Captions \label{Generate Video Captions, Instructions, and Edited Captions}}
We first use the video captions model to generate the caption for the given video. Based on the generated video captions, next we need to generate instructions and edited captions. This operation is solely within the text domain. For instance, given the input caption "a man in white is playing Tai Chi in the garden," we expect to generate instructions like "turn the man into a rabbit" and a modified caption "a rabbit is playing Tai Chi in the garden." The conduit for this creative endeavor is ChatGPT, owing to its immense knowledge and capability in human language. ChatGPT can generate creative yet sensible instructions and captions with just a few handwritten examples, as shown in Table \ref{tabel1}. For example, when presented with the video caption "a man in white is playing Tai Chi in the garden," the input prompt for ChatGPT is "please generate sentences as examples. Each sentence contains the original prompt 'a man in white is playing Tai Chi in the garden,' the instruction, and the modified prompt, which are separated by '$\vert$.'" After providing just two examples in practice, ChatGPT can generate instructions and edited captions as required.

\subsubsection{Generate Paired Videos \label{Generate Paired Videos}}
Next, we need to generate the edited video for the given video and instructions. Addressing this, we incorporate the Tune-a-Video pipeline \cite{RN10117}, a contemporary technique geared towards crafting temporally consistent videos through one-shot fine-tuning. By establishing correlations between captions and videos during fine-tuning, Tune-a-Video can subsequently generate a diverse array of temporally consistent edited videos. To enable this, we fine-tune the video diffusion model for each training clip, adhering to Tune-a-Video's configurations. After fine-tuning, the ChatGPT-generated edited captions serve as input to the Tune-a-Video model, resulting in the creation of temporally consistent videos that honor the original while reflecting the instructed changes. With the aforementioned pipeline, we could obtain a rich training dataset comprising input videos, instructions, and corresponding edited videos, forming the bedrock of the InstructVid2Vid model.

\subsection{InstructVid2Vid Model \label{InstructVid2Vid Model}}
In this section, we will introduce the model architecture of InstructVid2Vid. We will present the Inter-Frames Consistency Loss, meticulously devised to amplify the harmonious interplay between successive frames. Furthermore, we will introduce the multimodal classifier-free guidance to accommodate text-video input situations.

\subsubsection{Model Architecture \label{Model Architecture}}
\begin{figure*}
\centering
\includegraphics[width= \textwidth]{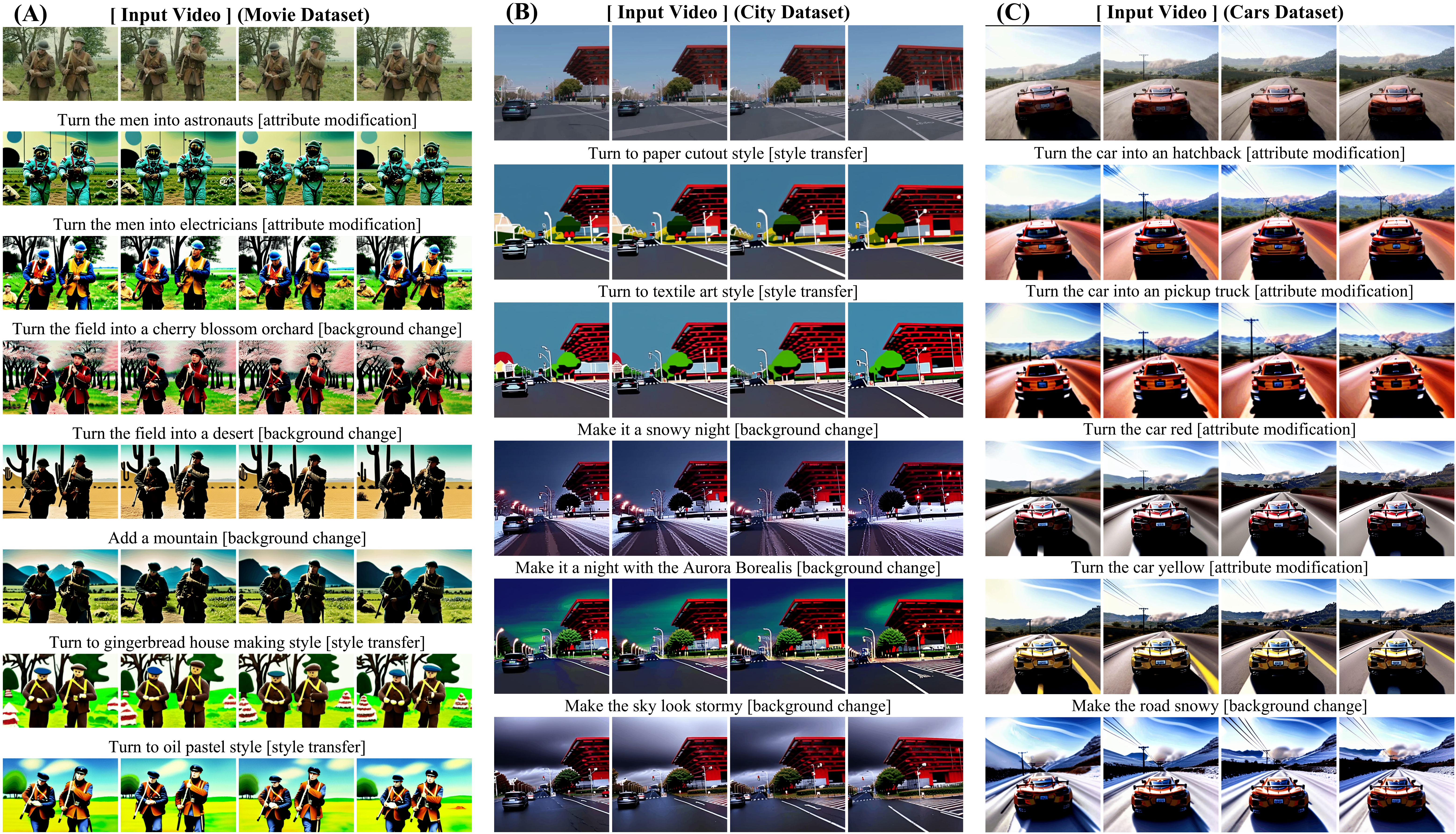}
\caption{Given the input video and text prompt, InstructVid2Vid could achieve attribute modification, background changes, and style transfer.}
\label{Figure2}
\end{figure*}

The InstructVid2Vid model is a video diffusion model, which contains VAE encoder, CLIP text encoder, and VAE decoder. The input of InstructVid2Vid consists of the input video and instruction. The video is encoded by the VAE encoder, which encodes every frame in the spatial domain. The instruction is encoded by the CLIP text encoder. The encoded video and instruction is then as input for the denoising network. After passing through the VAE decoder to reconstruct every video frame in the spatial domain, we can obtain the synthesized edited video.

The denoising network is inflated and modified from 2D U-Net \cite{RN10137} in T2I Stable Diffusion to suite video processing. We first modify the convolution in the 2D U-Net into a space-only 3D convolution. For example, we modify each $3\times3$ convolution into a $1\times3\times3$ convolution, where the first channel indexes video frames, and the second and third channels indicate the height and width of frames. After each spatial convolution, we add a temporal convolution to model the correlations in the temporal domain. Similar operations are also applied to the attention layer. We modify the attention block in the 2D U-Net into space-only attention, followed by attention in the temporal domain. The parameters in the spatial convolutions and attentions are inherited from the pretrained T2V stable diffusion for faster convergence speed, and the parameters in temporal convolutions and attentions are initialized randomly.

\begin{table*}[t]
\caption{Quantitative evaluation of the impact of Inter-Frames Consistency Loss on the smoothness and quality of the generated video. $\uparrow$ indicates higher values are preferable, while $\downarrow$ indicates lower values are preferable.}
\label{table2}
\begin{center}
\resizebox{\textwidth}{!}{
\begin{tabular}{cccc|cc}
\toprule
\multirow{2}{*}{Model} & \multicolumn{3}{c}{Consistency} & \multicolumn{2}{c}{Quality} \\
& Frame Differencing$\downarrow$&	Optical Flow$\downarrow$&	Block Matching$\uparrow$&	NR-PSNR$\uparrow$&	FID$\downarrow$ \\
\midrule
Vid2Vid-Zero & 40.55 & 14.91 & 4.05 & 42.15 & 228.48 \\
\midrule
InstructVid2Vid w/o Inter-Frames Consistency Loss & 38.91 & 12.86 & 4.63 & 46.34 & 104.36 \\
InstructVid2Vid with Inter-Frames Consistency Loss & 30.67 & 9.49 & 4.82 & 47.27 & 96.87 \\
\bottomrule
\end{tabular}}
\end{center}
\end{table*}

\subsubsection{Training with Inter-Frames Consistency Loss \label{Inter-Frames Consistency Loss}}
During training, the parameters in VAE encoder, CLIP text encoder, and VAE decoder are fixed, and we only optimize the denoising network. The optimization goal is to correctly predict the added noise while preserving the temporal consistency between adjacent frames. 

Distinct from the T2I diffusion model, the T2V model necessitates consistency between adjacent frames to ensure a smooth video. To imbue the InstructVid2Vid methodology with the capacity to engender fluidic videos, we propose Inter-Frames Consistency Loss—a novel augmentation seamlessly integrated into the training regime. The motivates the formulation of the Inter-Frames Consistency Loss, conceptualized as the divergence between anticipated characteristics of two successive frames, encapsulated in the following equation:
\begin{equation}
L_{fd}=\sum_{i=0}^{f}{v_i-v_{i-1}}
\label{Inter-Frames Consistency Loss Equation}
\end{equation}
where ${v}_i$ represents the encoded features of $i^{th}$ frame, and $f$ denotes number of frames. 

By incorporating the Inter-Frames Consistency Loss into the training process of the InstructVid2Vid model, we can guide the model to generate smooth edited videos that are consistent with the provided instructions and input videos. The final loss is defined as:
\begin{equation}
\begin{aligned}
L_{fd}&=L_{sd}+\lambda L_{fd} \\&
=\operatorname{MSE}\left(n_p,n_{gt}\right)+\lambda\sum_{i=0}^{f}{v_i-v_{i-1}}
\label{Inter-Frames Consistency Loss Equation2}
\end{aligned}
\end{equation}
\begin{equation}
v_i={\{{n}_{in}\}}_i-\{{n_p}\}_i
\label{Inter-Frames Consistency Loss Equation3}
\end{equation}
where $L_{sd}$ denotes the MSE loss used in Stable Diffusion. $n_p$ represents the predicted noise for the 3D U-Net, $n_{gt}$ refers to the ground truth noise level, and $n_{in}$ signifies the input noisy video. $\lambda$ is a hyperparameter defined during the training process. We set $\lambda=10^{-3}$ by default.

\subsubsection{Inference with Multimodal Classifier-Free Guidance \label{Multimodal Classifier-Free Guidance}}
During Inference, the InstructVid2Vid embraces a dual input paradigm comprising instructions and videos. This two-pronged dynamic necessitates a paradigm shift in classifier-free guidance, which must now dynamically allocate probability mass across both inputs. Envisioned as an innovation, the InstructVid2Vid model employs a dual-condition framework: a textual condition $c_T$ and a video condition $c_V$, harmoniously choreographing the synthesis of edited videos. 

The implementation of the classifier-free guidance mechanism involves the joint training of the diffusion model for both conditional and unconditional denoising. In the case of InstructVid2Vid, we randomly set $c_V=\emptyset$ for only 5\% of the training data, $c_T=\emptyset$ for another 5\% of the training data, and both $c_V=\emptyset$ and $c_T=\emptyset$ for the remaining 5\% of the data during training. By training the model using both conditional and unconditional diffusion, the InstructVid2Vid model can perform either conditional or unconditional denoising based on the text and video input.

During inference, the scores estimated for both conditional and unconditional denoising are combined to produce the final output. Given the text condition $c_T$ and video condition $c_V$, the following equation holds:
\begin{equation}
\begin{aligned}
&\log\left(P\left(z\mid c_T,c_V\right)\right) \\& =\log(\frac{P\left(z,c_T,c_V\right)}{P\left(c_T,c_V\right)}) \\ & =\log(\frac{P\left(c_T\mid c_V,z\right)P\left(c_V\mid z\right)P(z)}{P\left(c_T,c_V\right)})\\ &=\log\left(P\left(c_T\mid c_V,z\right)\right)+\log\left(P\left(c_V\mid z\right)\right) \\& +\log(P(z))-\log\left(P\left(c_T,c_V\right)\right)
\label{Classifier-Free Guidance Equation2}
\end{aligned}
\end{equation}

Taking the partial derivative with respect to $z$ yields:
\begin{equation}
\begin{aligned}
&\log\left(P\left(z\mid c_T,c_V\right)\right) \\ & =\nabla_z\log(P(z))+\nabla_z\log\left(P\left(c_V\mid z\right)\right) \\ & +\nabla_z\log\left(P\left(c_T\mid c_V,z\right)\right)
\label{Classifier-Free Guidance Equation3}
\end{aligned}
\end{equation}

This shifts probability mass towards higher values of $P\left(c_V\mid z\right)$ and $P\left(c_T\mid c_V,z\right)$. Consequently, during inference, the estimated noise at time step $t$ is calculated using the following formula:
\begin{equation}
\begin{aligned}
&\widetilde{e_\theta}\left(z_t,c_T,c_V\right)\\&=e_\theta\left(z_t,\emptyset,\emptyset\right)+s_V\cdot\left(e_\theta\left(z_t,c_V,\emptyset\right)-e_\theta\left(z_t,\emptyset,\emptyset\right)\right)\\ &+s_T\cdot\left(e_\theta\left(z_t,c_V,c_T\right)-e_\theta\left(z_t,c_V,\emptyset\right)\right)
\label{Classifier-Free Guidance Equation4}
\end{aligned}
\end{equation}
where $s_V$ and $s_T$ represent the video and text guidance scales.

\section{Experiments}
In this section, we illustrate the capability of the InstructVid2Vid to execute various video editing tasks, including attribute modification, background alteration, and style transfer, while maintaining temporal consistency across different types of videos. We will conduct an extensive quantitative and qualitative analysis to illustrate the superior performance of InstructVid2Vid.

\section{Evaluation Metrics \label{Evaluation Matrices}}
We utilize frame differencing, optical flow \cite{RN10164}, and block matching techniques to evaluate the consistency between adjacent frames in the edited video. The image quality of each video frame is assessed using NR-PSNR \cite{RN10162} and FID \cite{RN10163} metrics.

\textbf{Frame Differencing (FD)} is employed to determine the pixel differences between consecutive frames. The absolute value of the pixel differences is calculated for each pair of adjacent frames. A higher frame differencing value indicates that the video frames lack smoothness.

\textbf{Optical Flow (OF)} \cite{RN10164} is a method utilized to estimate pixel motion between adjacent frames. By computing the optical flow, a motion vector field can be obtained, which characterizes the motion information of objects in the scene. By comparing the motion vector fields of adjacent frames, the smoothness of the video frames can be evaluated. Significant changes in the motion vector field may suggest discontinuity between the video frames.

\textbf{Block Matching (BM)} divides each frame into multiple blocks, and subsequently, seeks matching blocks in adjacent frames. By calculating the degree of match and the motion vector between blocks, the smoothness of the video frames can be assessed. A lower degree of match or a larger motion vector may indicate poor smoothness between the video frames.

\textbf{No-Reference Peak Signal-to-Noise Ratio (NR-PSNR)} \cite{RN10162} is an objective quality assessment metric used for evaluating the generation of videos. Unlike traditional Peak Signal-to-Noise Ratio (PSNR), which necessitates a reference video for comparison, NR-PSNR can assess the quality of a video without such a reference. We employ a 3x3 Laplacian operator for edge detection, which highlights areas of rapid intensity change to estimate noise. The NR-PSNR is then calculated by taking the logarithm of the peak intensity-to-noise estimate ratio. A higher NR-PSNR value typically indicates better video quality for each video frame.

\textbf{Frechet Inception Distance (FID)} \cite{RN10163} is another measure employed to evaluate the quality of videos generated by machine learning models. FID captures the similarity between the frame distributions of the generated video and the real video by comparing their high-level features extracted using an Inception network \cite{RN129}. A lower FID score suggests a greater similarity between the distributions, indicating that the generated video is closer to the real video in the feature space.

\begin{figure}[t]
\centering
\includegraphics[width=0.48 \textwidth]{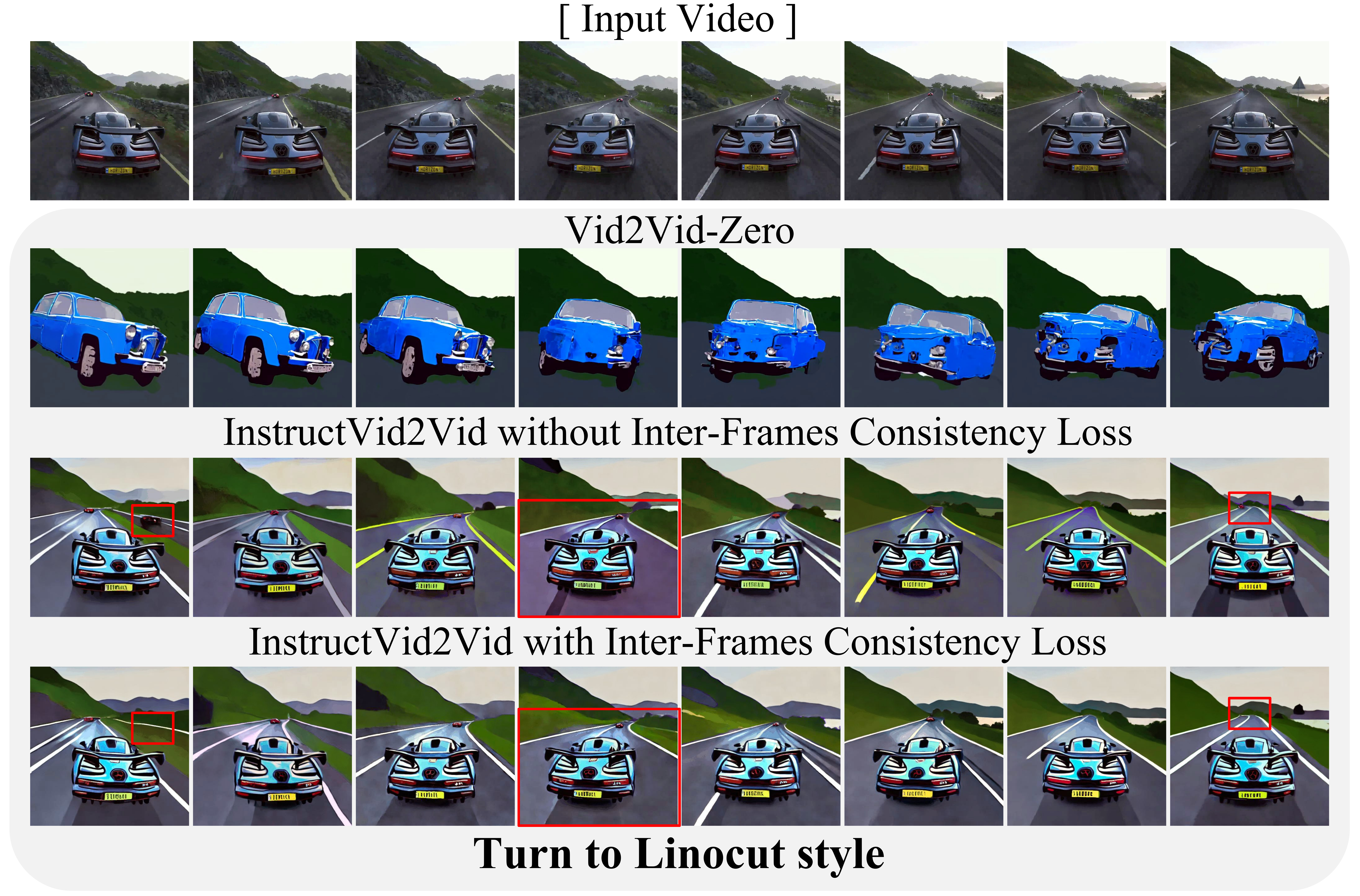}
\caption{Ablation Experiments of InstructVid2Vid. InstructVid2Vid with Inter-Frames Consistency Loss generates smoother videos compared to its ablation counterpart.}
\label{Ablation}
\end{figure}

\subsection{Visualization Results \label{Qualitative Evaluation}}
InstructVid2Vid effectively achieves attribute modification, background change, and style transfer in videos, while preserving temporal information and original motion, as shown in Figure \ref{Figure2}.

\textbf{Attribute Modification:} InstructVid2Vid can modify specific object attributes, like changing car colors (4th and 5th rows in Figure \ref{Figure2} (C)), while maintaining the original environment and motion. It can also alter appearances, such as soldiers' attire (2nd and 3rd rows in Figure \ref{Figure2} (A)), or transform vehicle types (2nd and 3rd rows in Figure \ref{Figure2} (C)).

\textbf{Background Change:} Seamlessly altering the background while preserving main subjects and temporal information, InstructVid2Vid can change scenes, e.g., soldiers walking in cherry blossoms (4th row in Figure \ref{Figure2} (A)) or create desert environments (5th row in Figure \ref{Figure2} (A)). It can adjust lighting, turning a city into a snowy night (4th row in Figure \ref{Figure2} (B)), and modify weather, like creating a thunderstorm (6th row in Figure \ref{Figure2} (B)).

\textbf{Style Transfer:} InstructVid2Vid achieves style transfer while preserving motion and semantics. For instance, it can transform a scene with soldiers into a gingerbread house or oil paste style (7th and 8th rows in Figure \ref{Figure2} (A)). It can also modify the style of city or car scenes, turning city views into a textile art style (3rd row in Figure \ref{Figure2} (B)).

\subsection{Quantitative and Qualitative Analysis \label{Ablation Experiments with Quantitative and Qualitative Analysis}}

In this section, we conduct a comprehensive analysis of InstructVid2Vid's performance, comparing it with the previous state-of-the-art Vid2Vid-Zero \cite{RN10176}. We also explore the impact of the proposed Inter-Frames Consistency Loss through ablation experiments.

For qualitative analysis, Figure \ref{Ablation} illustrates a comparison between InstructVid2Vid and Vid2Vid-Zero in video editing outcomes. InstructVid2Vid excels in synthesizing edited videos that closely match the source input, showing improvements in video quality with reduced non-conforming deformations. Notably, it exhibits enhanced frame consistency, resulting in greater similarity in shapes and appearances across adjacent frames.

Visualized ablation experiments, as shown in Figure \ref{Ablation}, compare videos generated by InstructVid2Vid with and without Inter-Frames Consistency Loss. Videos with the loss exhibit improved stability and fewer glitches, preventing errors seen in the absence of the loss, as depicted by the red rectangular regions in the comparison of the two ablation counterparts.

In the quantitative experiment, we evaluate generated videos using five indicators categorized into consistency among adjacent frames and video quality. InstructVid2Vid outperforms Vid2Vid-Zero in maintaining consistency among frames, as indicated by lower frame differencing and optical flow scores, and a higher block matching score. Additionally, InstructVid2Vid demonstrates higher video quality, evidenced by elevated NR-PSNR scores and lower FID values.

In the context of the Inter-Frames Consistency Loss ablation experiment, InstructVid2Vid with the loss exhibits higher frame consistency and superior video quality compared to the control group without the loss. This suggests that the Inter-Frames Consistency Loss enhances both the consistency between adjacent frames and the quality of individual frames by incorporating information from neighboring frames.

\section{Conclusion}
We introduce InstructVid2Vid, an innovative end-to-end diffusion-based video editing method guided by human language instructions. In contrast to prior approaches, InstructVid2Vid avoids specific fine-tuning or inversion for examples. Empirical results highlight InstructVid2Vid's proficiency in attribute modification, background changes, and style transfer while preserving object integrity and temporal coherence.

\bibliographystyle{IEEEtran}
\bibliography{my_endnote_cite, conference_cite}

\begin{thebibliography}{10}
\providecommand{\url}[1]{#1}
\csname url@samestyle\endcsname
\providecommand{\newblock}{\relax}
\providecommand{\bibinfo}[2]{#2}
\providecommand{\BIBentrySTDinterwordspacing}{\spaceskip=0pt\relax}
\providecommand{\BIBentryALTinterwordstretchfactor}{4}
\providecommand{\BIBentryALTinterwordspacing}{\spaceskip=\fontdimen2\font plus
\BIBentryALTinterwordstretchfactor\fontdimen3\font minus \fontdimen4\font\relax}
\providecommand{\BIBforeignlanguage}[2]{{%
\expandafter\ifx\csname l@#1\endcsname\relax
\typeout{** WARNING: IEEEtran.bst: No hyphenation pattern has been}%
\typeout{** loaded for the language `#1'. Using the pattern for}%
\typeout{** the default language instead.}%
\else
\language=\csname l@#1\endcsname
\fi
#2}}
\providecommand{\BIBdecl}{\relax}
\BIBdecl

\bibitem{RN10108}
T.~Brooks, A.~Holynski, and A.~A. Efros, ``Instructpix2pix: Learning to follow image editing instructions,'' \emph{arXiv e-prints}, p. arXiv:2211.09800, 2022.

\bibitem{RN10134}
E.~Molad, E.~Horwitz, D.~Valevski, A.~Rav~Acha, Y.~Matias, Y.~Pritch, Y.~Leviathan, and Y.~Hoshen, ``Dreamix: Video diffusion models are general video editors,'' \emph{arXiv e-prints}, p. arXiv:2302.01329, 2023.

\bibitem{RN10116}
S.~Liu, Y.~Zhang, W.~Li, Z.~Lin, and J.~Jia, ``Video-p2p: Video editing with cross-attention control,'' \emph{arXiv e-prints}, p. arXiv:2303.04761, 2023.

\bibitem{RN10117}
J.~Zhangjie~Wu, Y.~Ge, X.~Wang, W.~Lei, Y.~Gu, W.~Hsu, Y.~Shan, X.~Qie, and M.~Z. Shou, ``Tune-a-video: One-shot tuning of image diffusion models for text-to-video generation,'' \emph{arXiv e-prints}, p. arXiv:2212.11565, 2022.

\bibitem{RN10143}
R.~Zhang, X.~Hu, B.~Li, S.~Huang, H.~Deng, H.~Li, Y.~Qiao, and P.~Gao, ``Prompt, generate, then cache: Cascade of foundation models makes strong few-shot learners,'' \emph{arXiv e-prints}, p. arXiv:2303.02151, 2023.

\bibitem{RN10113}
J.~Li, D.~Li, C.~Xiong, and S.~Hoi, ``Blip: Bootstrapping language-image pre-training for unified vision-language understanding and generation,'' in \emph{Proceedings of the 39th International Conference on Machine Learning}, ser. Series BLIP: Bootstrapping Language-Image Pre-training for Unified Vision-Language Understanding and Generation, C.~Kamalika, J.~Stefanie, S.~Le, S.~Csaba, N.~Gang, and S.~Sivan, Eds., vol. 162.\hskip 1em plus 0.5em minus 0.4em\relax PMLR, 2022 Published, Conference Paper, pp. 12\,888--12\,900.

\bibitem{RN10176}
W.~Wang, K.~Xie, Z.~Liu, H.~Chen, Y.~Cao, X.~Wang, and C.~Shen, ``Zero-shot video editing using off-the-shelf image diffusion models,'' \emph{arXiv e-prints}, p. arXiv:2303.17599, 2023.

\bibitem{RN10179}
C.~Qi, X.~Cun, Y.~Zhang, C.~Lei, X.~Wang, Y.~Shan, and Q.~Chen, ``Fatezero: Fusing attentions for zero-shot text-based video editing,'' \emph{arXiv e-prints}, p. arXiv:2303.09535, 2023.

\bibitem{RN10180}
C.~Shin, H.~Kim, C.~H. Lee, S.-g. Lee, and S.~Yoon, ``Edit-a-video: Single video editing with object-aware consistency,'' \emph{arXiv e-prints}, p. arXiv:2303.07945, 2023.

\bibitem{RN10112}
O.~Bar-Tal, D.~Ofri-Amar, R.~Fridman, Y.~Kasten, and T.~Dekel, \emph{Text2LIVE: Text-Driven Layered Image and Video Editing}, ser. Lecture Notes in Computer Science, 2022, book section Chapter 41, pp. 707--723.

\bibitem{RN10148}
D.~Ceylan, C.-H.~P. Huang, and N.~J. Mitra, ``Pix2video: Video editing using image diffusion,'' \emph{arXiv e-prints}, p. arXiv:2303.12688, 2023.

\bibitem{song2021denoising}
\BIBentryALTinterwordspacing
J.~Song, C.~Meng, and S.~Ermon, ``Denoising diffusion implicit models,'' in \emph{International Conference on Learning Representations}, 2021. [Online]. Available: \url{https://openreview.net/forum?id=St1giarCHLP}
\BIBentrySTDinterwordspacing

\bibitem{RN10137}
O.~Ronneberger, P.~Fischer, and T.~Brox, \emph{U-Net: Convolutional Networks for Biomedical Image Segmentation}, ser. Lecture Notes in Computer Science, 2015, book section Chapter 28, pp. 234--241.

\bibitem{RN10164}
B.~K.~P. Horn and B.~G. Schunck, ``Determining optical flow,'' \emph{Artificial Intelligence}, vol.~17, no.~1, pp. 185--203, 1981.

\bibitem{RN10162}
D.~S. Turaga, Y.~Chen, and J.~Caviedes, ``No reference psnr estimation for compressed pictures,'' \emph{Signal Processing: Image Communication}, vol.~19, no.~2, pp. 173--184, 2004.

\bibitem{RN10163}
M.~Heusel, H.~Ramsauer, T.~Unterthiner, B.~Nessler, and S.~Hochreiter, ``Gans trained by a two time-scale update rule converge to a local nash equilibrium,'' in \emph{Proceedings of the 31st International Conference on Neural Information Processing Systems}, ser. Series GANs trained by a two time-scale update rule converge to a local nash equilibrium.\hskip 1em plus 0.5em minus 0.4em\relax Curran Associates Inc., 2017 Published, Conference Paper, p. 6629–6640.

\bibitem{RN129}
C.~Szegedy, V.~Vanhoucke, S.~Ioffe, J.~Shlens, and Z.~Wojna, ``Rethinking the inception architecture for computer vision,'' \emph{2016 Ieee Conference on Computer Vision and Pattern Recognition (Cvpr)}, pp. 2818--2826, 2016.

\end{thebibliography}

\end{document}